\documentclass[letterpaper, 10 pt, conference]{ieeeconf}  % Comment this line out if you need a4paper

\IEEEoverridecommandlockouts                              % This command is only needed if 
\usepackage{graphicx}
\usepackage{float}
\usepackage{amsmath}
\usepackage[section]{placeins}
\usepackage{lipsum} 
\usepackage{environ} 
\usepackage{amsmath}
\usepackage{tabularx}
\usepackage{adjustbox}
\usepackage[dvipsnames]{xcolor}
\usepackage{comment} 
\usepackage{setspace}
\usepackage{fixltx2e}
\usepackage{balance}

\title{\LARGE \bf
Employing Universal Voting Schemes for Improved Visual Place Recognition Performance
}

\author{Maria Waheed$^{1}$,  Michael Milford$^{2}$, Xiaojun Zhai$^{1}$, Maria Fasli$^{1}$, Klaus McDonald-Maier$^{1}$ and Shoaib Ehsan$^{1,3}$%  <-this % stops a space
\thanks{*This research was supported by the UK Engineering and Physical Sciences Research Council (EPSRC) through grants EP/R02572X/1,  EP/P017487/1 and EP/V000462/1. This research was also partially supported by funding from ARC Laureate Fellowship FL210100156 to MM and the QUT Centre for Robotics. \textit{(Corresponding author: Maria Waheed)}}% <-this % stops a space
\thanks{$^{1}$M. Waheed, X. Zhai, Maria Fasli, K. McDonald-Maier and S. Ehsan are with the School of Computer Science and Electronic Engineering, University of Essex, Colchester CO4 3SQ, United Kingdom.
        {\tt\small (e-mail: mw20987@essex.ac.uk; xzhai@essex.ac.uk; mfasli@essex.ac.uk; kdm@essex.ac.uk; sehsan@essex.ac.uk;  )}}%
\thanks{$^{2}$M. Milford is with the School of Electrical Engineering and Computer
Science, Queensland University of Technology, Brisbane, QLD 4000, Australia
        {\tt\small (e-mail: michael.milford@qut.edu.au)}}%
\thanks{$^{3}$Shoaib Ehsan is also with the School of Electronics and Computer Science, University of Southampton, Southampton, SO17 1BJ
        {\tt\small (e-mail: s.ehsan@soton.ac.uk)}}        
}

\begin{document}

\maketitle
\thispagestyle{empty}
\pagestyle{empty}

%%%%%%%%%%%%%%%%%%%%%%%%%%%%%%%%%%%%%%%%%%%%%%%%%%%%%%%%%%%%%%%%%%%%%%%%%%%%%%%%
\begin{abstract}

Visual Place Recognition has been the subject of many endeavours utilizing different ensemble approaches to improve VPR performance. Ideas like multi-process fusion, Fly-Inspired Voting Units, SwitchHit or Switch-Fuse involve combining different VPR techniques together, utilizing different strategies. However, a major aspect often common to many of these strategies is voting. Voting is an extremely relevant topic to explore in terms of its application and significance for any ensemble VPR setup. This paper analyses several voting schemes to maximise the place detection accuracy of a VPR ensemble set up and determine the optimal voting schemes for selection. We take inspiration from a variety of voting schemes that are widely employed in fields such as politics and sociology and it is evident via empirical data that the selection of the voting method influences the results drastically. The paper tests a wide variety of voting schemes to present the improvement in the VPR results for several data sets. We aim to determine whether a single optimal voting scheme exists or, much like in other fields of research, the selection of a voting technique is relative to its application and environment. We propose a ranking of these different voting methods from best to worst which allows for better selection. While presenting our results in terms of voting method's performance bounds, in form of radar charts, PR curves to showcase the difference in performance and a comparison methodology using a McNemar test variant to determine the statistical significance of the differences. This test is performed to further confirm the reliability of outcomes and draw comparisons for better and informed selection a voting technique.

\end{abstract}

%%%%%%%%%%%%%%%%%%%%%%%%%%%%%%%%%%%%%%%%%%%%%%%%%%%%%%%%%%%%%%%%%%%%%%%%%%%%%%%%

\section{INTRODUCTION}
Efficiently performing the task of visual place recognition still remains an open field of research since this apparently simple task of recognizing a previously visited place is\begin{figure}[h!]
    \centering
    \vspace*{0.1in}
    \includegraphics[width=1\columnwidth]{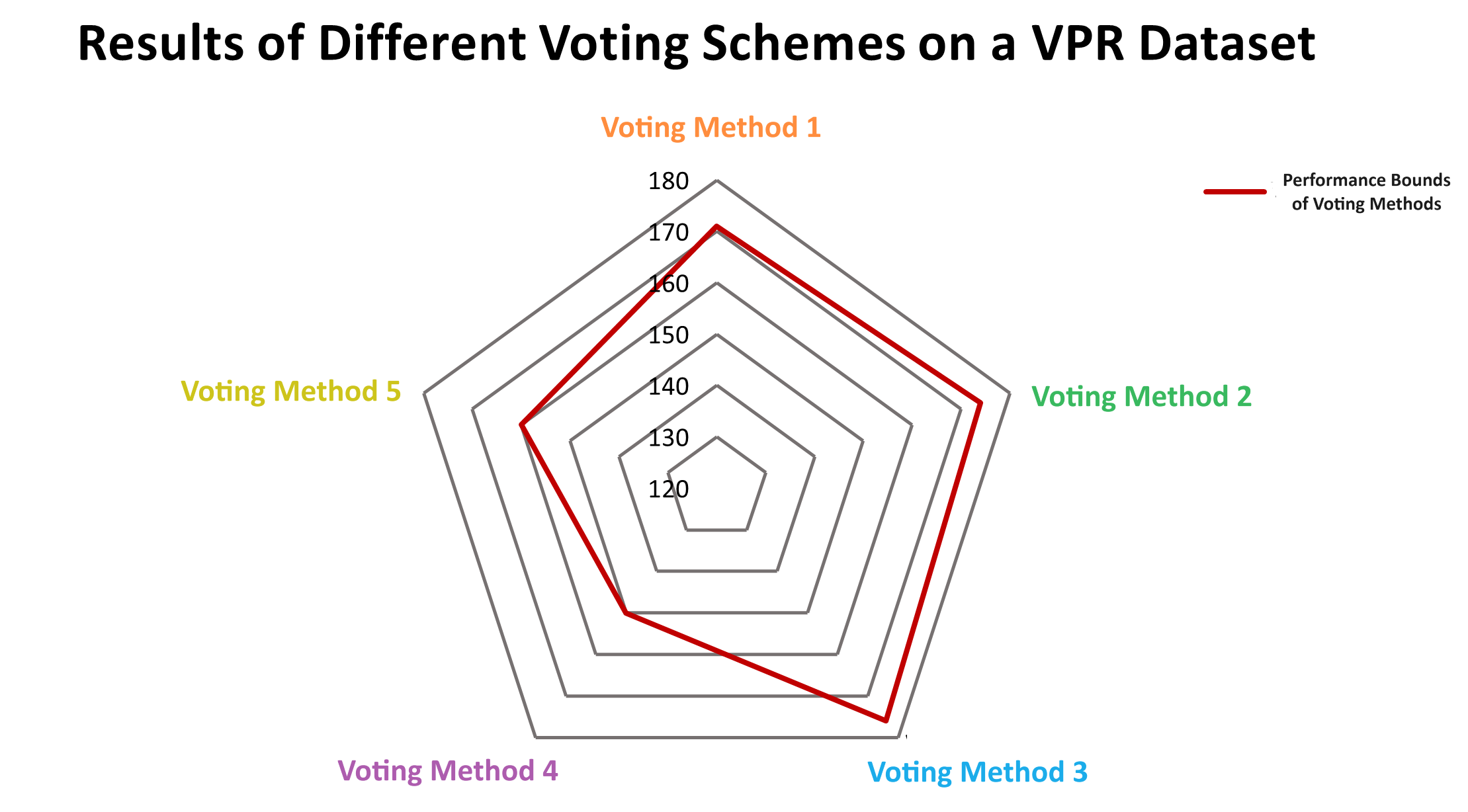}
    \caption{Sample output from the proposed experimental setup to evaluate the difference in performance bounds between each voting methodology. The red line in the radar chart is representative of the performance bounds (images correctly matched) and the axis represent the total number of query images in the data set. The various dimensions are useful to interpret the difference in performance of a voting scheme, in comparison to the others, such that the closer the red line is to boundary the better the performance of the voting scheme.   }
    \label{fig:my_label}
\end{figure} quite complex due to the extremely varying nature of the environments a robot can encounter. From low to severe viewpoint, illumination or even seasonal variations highly contribute to the difficult nature of the problem [1]-[3]. Although, many excellent VPR techniques are available to the robotics community to tackle this task, there is no universal VPR technique that can perform equally well in all types of variations encountered. Rather than another attempt to develop a new VPR technique from scratch, other intuitive solutions such as [4],[5] that introduce the concept of multi-process fusion between different VPR techniques. Furthermore, to refine the idea of multi-process fusion, [6] proposes a framework evaluating the complementarity of different VPR techniques, while SwitchHit [7] presents combining abilities of different VPR techniques by merely switching to a better technique based on complementarity and [8][9] are other relevant examples of ensemble based methods for VPR. The common phenomena among all these ensemble approaches is that they use some form of voting. Voting or voting classifiers in the most conventional sense when used to solve machine learning problems have very limited options, hard and soft voting. While on the other hand, voting as a concept when employed universally offers a pool of options each with its own set of individual
\begin{comment}
\begin{table*}[h!]
\centering
\caption{Voting Schemes and Their Properties}
\begin{tabular}{|c|c|c|c|c|c|}
\hline
\textbf{Voting Procedure} & \textbf{Type of   Voting}  & \textbf{Consistent} & \textbf{Independent} & \multicolumn{1}{c|}{\textbf{Neutral}} \\ \hline
Condorcet Voting          & Single winner/majoritarian & No                  & No                   & Yes                                   \\ \hline
Contingent Voting         & Single winner/majoritarian & Yes                 & No                   & Yes                                   \\ \hline
IRV Voting                & Multi-stage/Positional     & No                  & Yes                  & Yes                                   \\ \hline
Broda Count Voting        & Positional                 & Yes                 & No                   & Yes                                   \\ \hline
Plurality Voting          & Positional                 & Yes                 & Yes                  & Yes                                   \\ \hline
\end{tabular}
\end{table*} 
\end{comment}
characteristics. Furthermore, each voting scheme due to its unique methodology produces different results hence it is correct to assume that selection of a voting scheme in any field is not a trivial task. Researchers in different fields have made the effort to tackle with the lack of standardization among voting schemes and determine the optimal voting schemes for different types of tasks. However, an attempt to employ universal voting methods or an exploration to determine which of the voting schemes is optimal to use in an ensemble VPR set up is an area that has not yet been attempted. 

We present and evaluate the applications of universally employed voting schemes on a standard VPR ensemble set up with multiple VPR techniques to observe the difference in performance and results for a varied set of VPR data sets. A  variety of voting systems that have unique characteristics and are widely employed in other fields such as politics and sociology have been selected for testing. We attempt to conclude whether a single best voting method exists from among these voting scheme or much like other fields it is case specific. We presents our results in terms of performance bounds of voting schemes on different datasets much like in Figure. 1 , PR curves for comparing accuracy and determine the statistical significance of the performance difference via McNemar's test and Z-scores values.

The rest of this paper is organized as follows. Section II provides an overview of related work. Section III presents the methodology explaining the  different voting schemes and their method of employment. Section IV describes the experimental setup. The results based on testing these different voting schemes are presented in Section V and finally, conclusions are given in section VI.

\section{Related Work}
This section provides an overview of the related work on the challenges in the field of visual place recognition and different methods proposed to efficiently solve the problem of VPR. After exhaustive research for developing new VPR techniques from hand-crafted methods such as GIST [10,12] and the bag-of-visual-words [11]  to Convolutional Neural Networks (CNNs) such as NetVLAD [13] and RegionVLAD [14]. 

The research has recently shifted towards searching for other innovative solutions which has led to the development of ensemble VPR methods [4]-[9]. One such approach is multi-process fusion where [4],[5] work by selecting and combing different VPR  techniques to  match the query image from a sequence of images generated. The authors of [4] combined multiple image processing methods to decide the best match from the sequence of images generated. While the multi-process fusion system introduced in [5]  combines multiple VPR methods using a Hidden Markov Model (HMM) and presents  a three-tier hierarchical multi-process fusion system which is customizable and may be extended to any arbitrary number of tiers. Another interesting idea that explores the notion of complementarity between multiple VPR techniques is introduced by [6]. A McNemar’s test like approach is used to test out the level of complementarity between different VPR techniques. Building on the work of complementarity between multiple VPR techniques SwitchHit [7] proposes a complementarity based systems between different VPR techniques  which switches and selects between different VPR techniques based on probability of correct match and complementarity. Moreover, other interesting work is also presented by [8],[9] where [8] attempts to gather a collection of very small CNN voting units to improve VPR performance. This is achieved by selecting the prediction of matched image that most units agree on. [9] discusses the idea of probabilistic voting for VPR utilizing nearest neighbor descriptor voting.

Finally it is also important to discuss that the two conventionally employed voting classifiers for most of these discussed ensemble methods are either some variants of hard or soft voting. Hard voting which is quite similar to the basic majority voting style and simply selects the the candidate with the highest number of votes while soft voting instead gathers the probabilities and selects the candidate with highest probability.

\section{Methodology}
This section presents the different voting schemes and their methodologies, that have been employed to a series of different VPR data sets to observe how the use of different voting mechanisms effects overall results in a basic ensemble VPR system. We test these voting schemes in a VPR set up that is simultaneously employing all state-of-the-art VPR techniques available and the final step involves selecting the correct reference image by using a voting methodology. The total number of VPR technqiues for this experiment are the \textit{voters} while all reference images are the \textit{candidates} and the  selected top reference image/images by each VPR technique represents their \textit{votes}.

The structure of our methodology section is based on analyzing different voting schemes that have been carefully selected to include unique voting systems to be tested. This work will help determine whether there is a clear winner when selecting the type of voting technique to employ or if its a relative choice dependant on other factors for example dataset type.

\subsection{Voting Scheme I : Plurality Voting}

This type of a voting mechanisms belongs to the family of positional voting which is a voting system that involves different ranks for different candidates and each rank holds a different priority. For plurality voting the \textit{candidate} which has the most first place \textit{votes} than any other \textit{candidate} in the run is selected to be the final match. This is the most common or basic type of voting and is quite similar to hard voting that is used when trying to solve a classification problem. When dealing with an ensemble of VPR techniques each of which results in different reference images selected to match with the query image, it is not always simple or obvious which image is the correct match to the query. Here we present how plurality voting is employed to test out the results it would produce over different VPR data sets.

Let $c$ be the retrieved image and $v\textsubscript{c}$  is the number of votes each retrieved image acquires. $Argmax$ returns the reference image with the maximum votes, which is the final selected matched image to the query.

\begin{equation}
    \centering
    argmax \; c\in \{1,2,...,n\} \; v_{c}
\end{equation}

\subsection{Voting Scheme II : Condorcet Voting}
Condorcet is a ranked type of voting method that attempts to determine the overall selection of the \textit{candidate}, reference image, by comparing the results of each \textit{voter}, VPR technique, from the ensemble techniques in one-on-one match-ups. Each VPR technique produces a sequence of potential correct images that are ranked on the basis of matching scores and a matrix is created with the number of times each image is ranked higher than the other possible correct matches. This matrix represents the \textit(ranked votes), selected reference image, and is used to determine the Condorcet winner. The reference image to win the most head-to-head match-ups is used to\begin{figure} [tb]
    \centering
    \vspace*{1mm}
    \includegraphics[width=1\columnwidth]{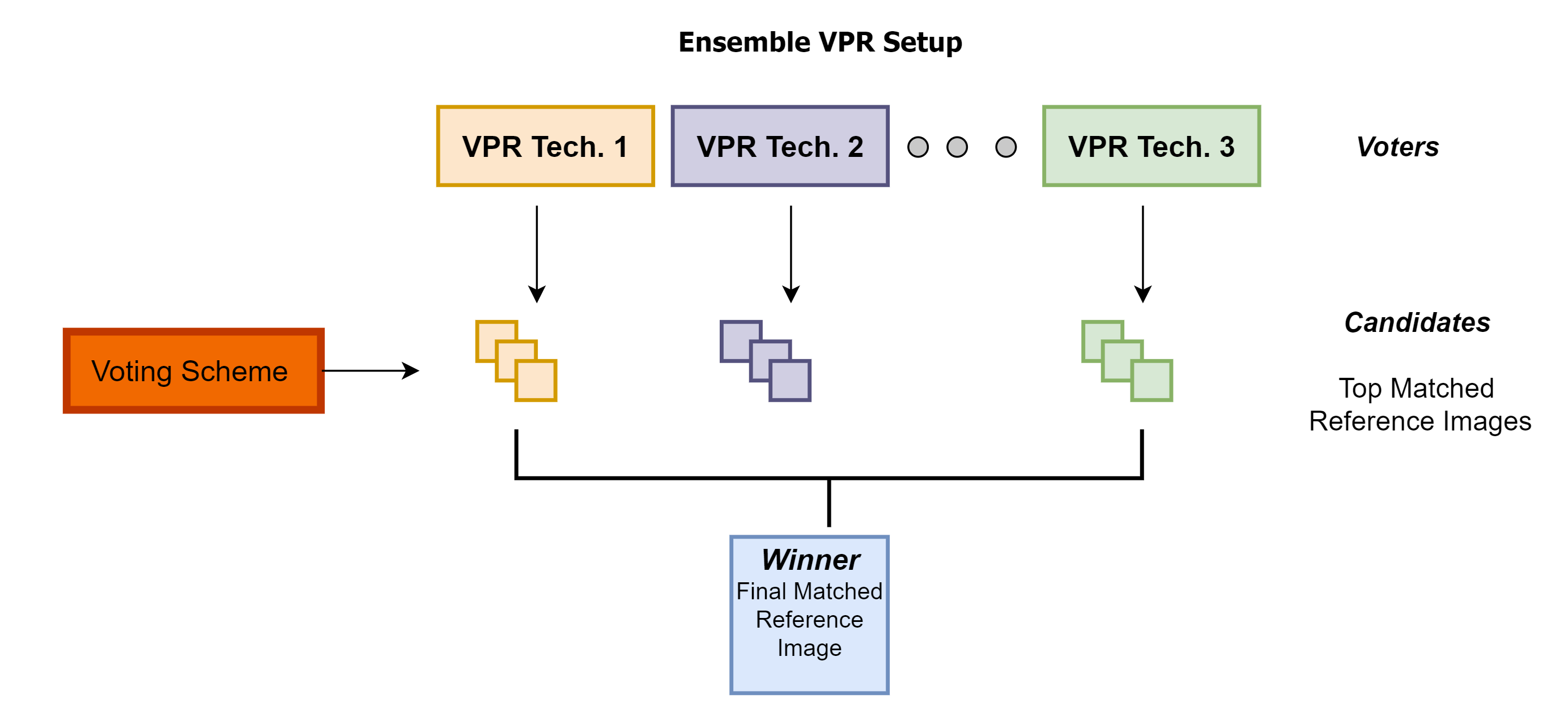}
    \caption{A standard VPR ensemble set up employing several VPR methods simultaneously, that produces the top best matches by each method which are subjected to various voting schemes to observe difference in results.}
    \label{fig:my_label}
\end{figure}\begin{table}[]
\centering
\caption{VPR-Bench Datasets [36]} 
\begin{tabular}{|l|c|c|c|}
\hline
\textbf{Dataset}      & \multicolumn{1}{l|}{\textbf{Environment}} & \multicolumn{1}{l|}{\textbf{Query }} & \multicolumn{1}{l|}{\textbf{Ref. Images }} \\ \hline
GardensPoint & University                          & 200                                        & 200                                      \\ \hline
ESSEX3IN1   & University                          & 210                                        & 210                                      \\ \hline
CrossSeasons & City-Like                                 & 191                                        & 191                                      \\ \hline
Corridor     & Indoor                                    & 111                                        & 111                                      \\ \hline
17Places    & Indoor                                    & 406                                        & 406                                      \\ \hline
Livingroom   & Indoor                                    & 32                                         & 32                                       \\ \hline
\end{tabular}
\end{table} determine the \textit{winner}, matched reference image, based on the strength of their victories in the pairwise match-ups.

Let C be the set of references images where n is the total number of reference images. 
\begin{equation}
    \centering
     C = \{c_1, c_2, ..., c_n\}
\end{equation}

Let's denote the ranked positions as B where v is the total number of ranks. 
\begin{equation}
    \centering
     B = \{b_1, b_2, ..., b_v\}
\end{equation}

\begin{figure*}[!htb]
    \vspace*{0.1in}
    \includegraphics[width=2\columnwidth]{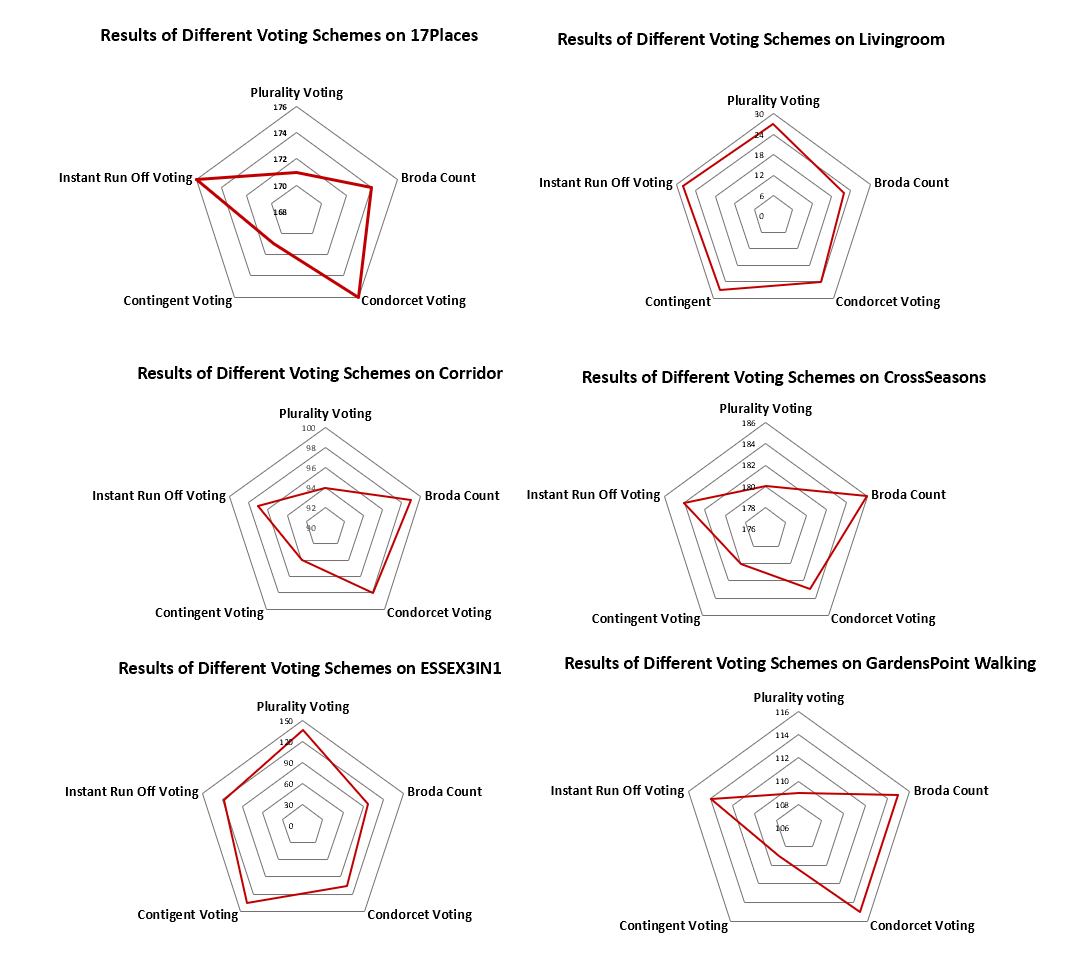}
   \caption{Difference in performance bounds of each voting methodology including Plurality, Condorcet, Contingent, Broda Count and Instant Run Off voting, in terms of query images correctly matched for different data sets : 17Places (top left), Livingroom (top right), Corridor (center left), CrossSeasons (center right), ESSEX3IN1 (bottom left) and GardensPoint (bottom right).  
   }
    \label{figurelabel}
\end{figure*}

Each \(b\textsubscript{v}\) in \(B\) contains a ranked preference order for the reference images in C. For example, if we have 3 references images \(c\textsubscript{1}\),\(c\textsubscript{2}\),\(c\textsubscript{3}\) a ranked positions could be \(b\textsubscript{v} (c\textsubscript{1},c\textsubscript{2},c\textsubscript{3})\), indicating that \(c\textsubscript{1}\) is the technique’s first choice, \(c\textsubscript{2}\) is the second choice, and \(c\textsubscript{3}\) is the third choice. The Condorcet winner is the reference image that would win in a head-to-head matchup against any other candidate. To calculate the pairwise victories for each pair of candidates \(c\textsubscript{i}\) and \(c\textsubscript{j}\) in C, we add the number of times \(c\textsubscript{i}\) is ranked higher than \(c\textsubscript{j}\). The reference image that is ranked higher than all other images is the final selected reference image. 
\\
\begin{equation}
    \centering
\sum{c\textsubscript{i} > c\textsubscript{j}}
\end{equation}

\subsection{Voting Scheme III : Broda Count Voting}
Broda Count is another positional voting system utilized for various types of electoral tasks. Similar to the plurality voting it also belongs to the family of positional voting where the \textit{candidates}, potential matches, to the query image are ranked in a descending order based on their matching scores. The position or rank of the reference image is important as a higher rank (i.e higher points/score) suggests a higher chance or preference of the particular image being selected as the final match in the ensemble VPR set up.

Let \(c\) be the reference image selected while \(i\) represents the total number of techniques being considered. \(j\) represents the rank of each reference image and $n$ is the value of points or score for a given rank. And finally \(S\textsubscript{c}\) is the summation of all the points for a single reference image. 
\begin{equation}
    \centering
    S_{c} = \sum_{i=1}^{i=n} x_{i,j}
\end{equation}
The image with the highest sum is selected as the final match to the query. 
\begin{equation}
    \centering
    argmax \; i\in \{1,2,...,n\} \; S_{c}
\end{equation}

\subsection{Voting Scheme IV : Contingent Voting  }

Contingent Voting which is a form of rank-choice voting in which the \textit{candidates}, references images, are ranked while each rank represents a different priority. This type of voting on an ensemble VPR set up allows to ensure that the final selected image has the broadest possible support/votes from among all the VPR techniques being employed. 

In a contingent voting system the first-choice votes meaning the top most candidates/images with the highest first place votes in terms of ranks are tallied. If the a reference image has an absolute majority (more than 50\%) of the first-choice votes it is simply selected as the final image. If, however, no reference image has an absolute majority, the images with the lowest votes are eliminated and their ranks and scores are transferred to the image with highest votes. Lastly, the steps for recount and redistribution are performed by taking into account the recounted votes and the process is repeated until a single best winner for a reference image is found. Although there are several other ways for counting and redistributing the votes for a contingent system but \begin{figure*}[h!]
    \vspace*{0.1in}
    \includegraphics[width=2\columnwidth]{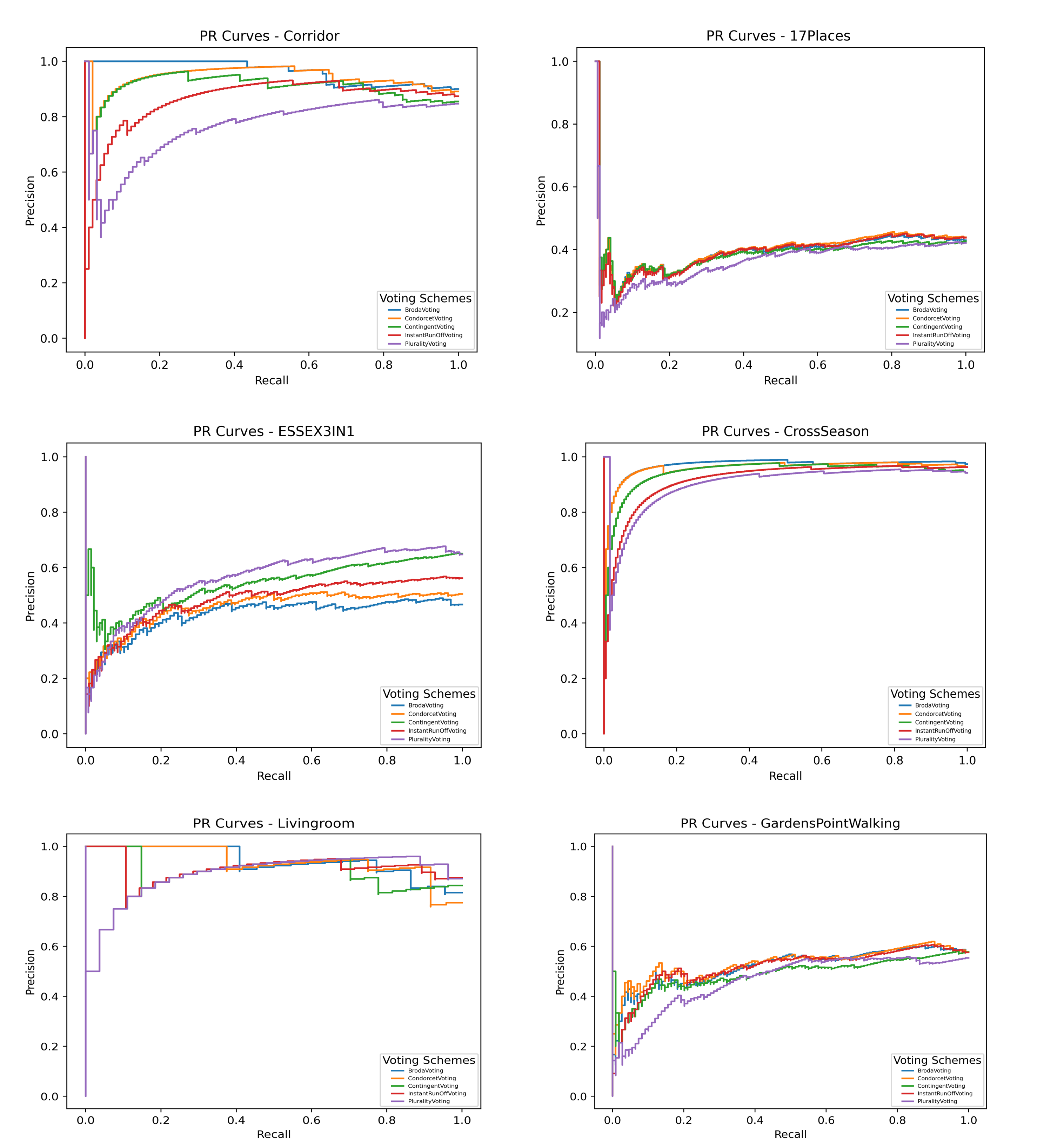}
   \caption{Performance in terms of PR curves for each voting methodology including Plurality, Condorcet, Contingent Voting, Broda Count and Instant Run Off voting, in terms of query images correctly matched for different data sets : 17Places (top left), Livingroom (top right), Corridor (center left), CrossSeasons (center right), ESSEX3IN1 (bottom left) and GardensPoint (bottom right).  
   }
    \label{figurelabel}
\end{figure*} the core principle is the same one as employed in this, where the VPR techniques select the possible reference images matches and rank their preferences and the voting aims to select a reference image that has the broader votes/support  of selection among all the employed VPR methods. 

Contingent voting doesn't lend itself to a single mathematical formula like some other voting systems, but it involves a series of steps that can be explained mathematically. Each VPR method employed in an ensemble VPR set up puts forth a ranked ballot of reference images to match and these \begin{figure*}[h!]
    \vspace*{0.1in}
    \includegraphics[width=2\columnwidth]{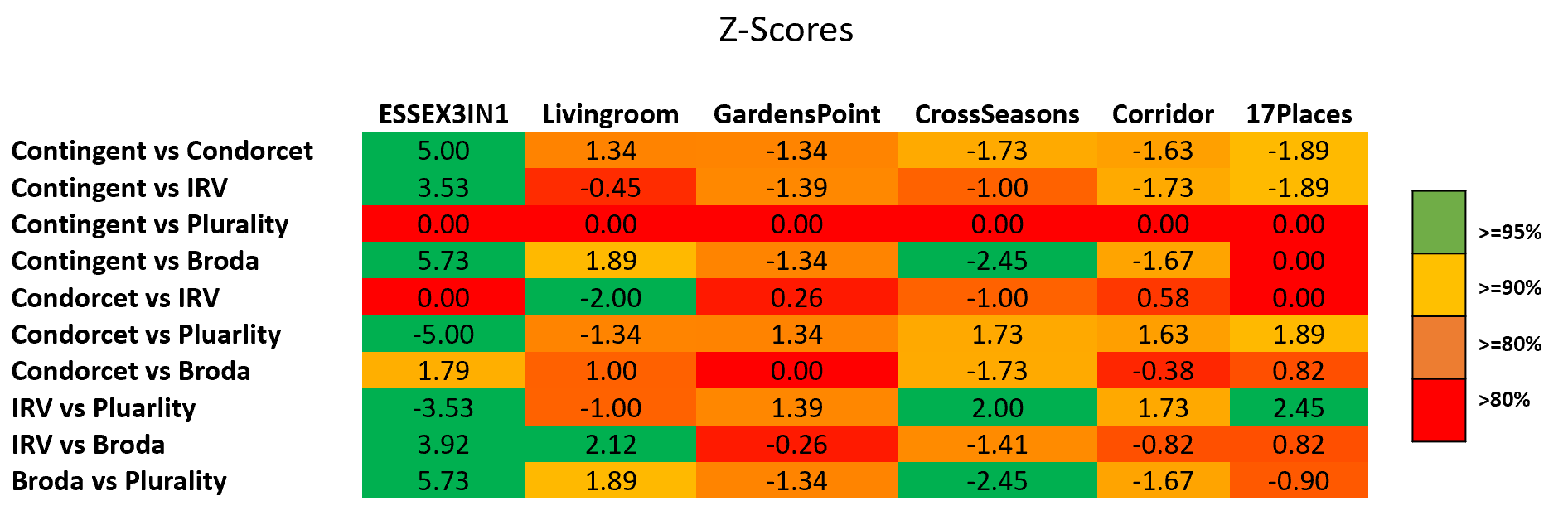}
   \caption{Pairwise comparisons between the voting methods have been considered. A sign convention is used to present the results: a positive value of Z indicates that the first method of the pair outperforms the second one, whereas a negative Z score has the opposite meaning. Corresponding to the z-scores the legend represents the confidence intervals starting with the highest confidence intervals being green to lowest being red.
   }
    \label{figurelabel}
\end{figure*} are assigned preferences such as "1" for the first choice, "2" for the second choice, and so on.

For representation, let \(c\) be the set of references images where \(n\) is the total number of reference images. 
\begin{equation}
    \centering
     C = \{c_1, c_2, ..., c_n\}
\end{equation}

Let's denote the ranked positions as \(B\) where \(v\) is the total number of ranks.

%\begin{figure*}[!htb]
 %   \vspace*{0.1in}
  %  \includegraphics[width=2\columnwidth]{ExtendedVotingBarcharts.png}
   %\caption{Performance percentage difference between each voting methodology inincluding Plurality, Condorcet and Broda Count voting, in terms of query images correctly matched for different data sets : Corridor (top left), Livingroom (top right), 17Places (center left), ESSEX3IN1 (center right), GardensPoint Walking (bottom left) and CrossSeasons (bottom right). 
   %}
    %\label{figurelabel}
%\end{figure*}

\begin{equation}
    \centering
     B = \{b_1, b_2, ..., b_v\}
\end{equation}
For counting first-choice votes, let \(N\) be the total number of VPR techniques and let V\textsubscript{i} be the no. of first choice votes recorded by the candidate i. For checking for an absolute majority, if any reference image receives more than 50\% of (N/2) of the first-choice votes, that reference is the final selected the image to match. To explain if a reference image d exists such that \(V(d)\) is greater than \(N/2\), then the reference image is selected. 

For elimination of the lowest ranked image, the image with the least first choice votes is eliminated while their votes are redistributed to the highest ranked images. 
%Let \(C\) be the image with the least first-choice votes and let \(R(c)\) be the set of techniques who ranked c as their first choice. For each VPR technique \(v\) in \(R(c)\), their vote is transferred to the image that is their next highest-ranked preference for the reference image. 
To recount and recalculate, the votes are recounted with the redistributed votes and the process is repeated to check for the absolute majority winner until an image receives more than 50\% of the votes or until only one image remains to be selected.

\subsection{Voting Scheme V : Instant Run Off Voting  }
Instant Runoff Voting (IRV) is another type of ranked voting used to ensure the selection of a candidate with the broadest support/votes is selected. When selected reference images in order of preference, in this case in terms of highest to lowest similarity scores. Then for tabulation of the votes, if any reference image receives an absolute majority it is simply selected as the final reference image, similar to Contingent voting. If, however, no such reference image exists the votes are recounted and redistributed to their second-choice candidates. After redistribution the votes are recounted, and the process repeated until a candidate with a majority is selected. 

Let \(N\) be the number of VPR techniques being employed and let \(C\) be the set of reference images where \(n\) is the total of images. While each \(N\) ranks all reference images a from \(1\) to \(n\). 
\begin{equation}
    \centering
     C = \{c_1, c_2, ..., c_n\}
\end{equation}

The next step is to find the first-choice votes for each image and determine whether an absolute winner exists, which if it does the process is stopped and a final image is selected. If, however no such winner is found identify the candidate with the fewest first-choice votes, let’s call this reference image \(D\). Reference image \(D\) is then eliminated and all votes for \(D\) are redistributed to the next-highest-ranked reference image. This process ensures that each vote is transferred to the next available preference.
The votes after this redistribution are recounted without the eliminated reference images, to determine an image with an absolute majority. And the process continues until a reference image achieves a majority.

\section{Experimental Setup}

The evaluation of these different voting methodologies help observe how a voting scheme can have significantly different results than another for the same type of ensemble VPR set up. For this experiment different VPR data sets have been utilzied to test this idea and are presented in TABLE II lists all the data sets including Corridor [16], Living room [15], ESSEX3IN1 [17], GardensPoint [18], Cross-Seasons [19], and 17Places [20], along with their environmental set up, number of query and reference images. Eight state-of-the-art VPR techniques have been selected for the experimentation, to test the different voting schemes on an ensemble set up and these include AMOSNet [21], HOG [22],[23], and  AlexNet [24],[25], HybridNet [26], NetVLAD [14], RegionVLAD [15], CoHOG [27], and CALC [28].

\section{Results} 
In this section we present the results produced by testing each of the selected voting methodologies over various VPR datasets, under the same ensemble VPR set up. The results are presents in three different categories to efficiently evaluate the utility of each voting method. Firstly, the results are presented in terms of the performance bounds of each Voting method, PR curves for accuracy and z-score for testing significance in performance difference.

Figure. 3 presents the results that were produced starting from 17Places data set for which two of the voting methods, Instant Run off and Condorcet Voting have the highest performance bounds followed by Broda Count and then Plurality and Contingent Voting methods. For the Livingroom dataset an overall uniform performance bound is observed for most methods but with plurality, Instant run off and Contingent voting slightly outranking the others. The next two data sets are Corridor and CrossSeasons, for both of which Broda Count ranks the highest among all methods followed by Contingent Voting and then the others. The last two data sets tested for their performance bounds are ESSEX3IN1 and GardensPoint where ESSEX3IN1 has similar performance bounds for most methods although Contingent Voting outranks the others slightly. GardensPointWalking has the least favourable performance bounds for Contingent and Plurality Voting while Condorcet, Broda and Instant Run Off have better performance bounds overall. 

Next, in Figure. 4 we present our results in terms of PR curves to show case the performance difference observed for the different voting methods tested for multiple VPR data sets. A varied results in performance is observed beginning from the Corridor Data set where Instant Run Off and Condorcet voting outperform the remaining three substantially. The livingroom dataset has a more uniform performance in terms of the precision-recall observed over this dataset with most voting methods performing similar. The Corridor and CrossSeasons dataset both have the highest performance when utilizing Broda Count as the voting method followed by Condorcet Voting. For the ESSEX3IN1 dataset plurality and Contingent voting appear to produce better accuracy. While, GardensPoint data has the opposite results as it has least favourable results when employing plurality and Contingent Voting. However, Condorcet Voting appears to produce better results in comparison to other voting methods tested. 

\subsection{McNemar's Test to Determine Performance Difference Significance:}
To further confirm our results presented in Figure 3 and 4, our not a chance occurrence, rather conclusive evidence of the significant difference between different voting schemes, we utilize a variant of the McNemar’s test [29]. This will help identify the statistically relevant performance differences, with a confidence interval, between the various voting schemes and in turn distinguishing the best cases from worst cases. We adapt the approach described in [29] to perform a pairwise evaluation of voting methods using a series of frame-by-frame matches/mismatches on the same data set. Since the McNemar’s test [30, 31] cannot be used to compare more than two VPR methods simultaneously, so a series of independent pairwise tests are necessary to compare multiple voting methods. We present these results in the form of a Z-Score table that states the different Z-scores corresponding to their confidence intervals to showcase the statistical significance. In particular, 95\% significance level corresponds to Z = 1.96 and presents a highly significant performance difference between the two compared voting schemes. 
\begin{equation}
    \centering
     X^{2} =\frac{\left |N_{sf} -N_{fs} \right |}{\sqrt{N_{sf}+N_{fs}}}
\end{equation}

The value to generate the Z-score utilizing the two-tails table is generated as $N\textsubscript{sf}$ denotes the number of trials where the voting 1 succeeded, and 2 failed; $N\textsubscript{fs}$ denotes the number of trials where 1 failed and 2 succeeded. $X\textsubscript{2}$ is distributed, to a good approximation, as chi-squared with one degree of freedom. Where the confidence interval associated with Z can be determined using tables [190]. The results are presented in Figure. 5 in form of a heat-map to efficiently locate pairs and voting methods that are significantly better than their compared counterparts. The table can be interpreted using the Z scores mentioned and the corresponding confidence interval to understand whether the difference is significant or not. 

Figure. 5 presents results in terms of different data sets tested such as ESSEX3IN1 for which overall Contingent voting scheme is significantly better in performance than the others. The second best option for selection can be concluded as IRV over Plurality or Broda and lastly Broda is a better selection over Plurality voting. Again, the results demonstrate how substantial the difference can be in selection if voting schemes are considered based on performance rather than simply selecting the conventional approach. Considering the results for Livingroom Dataset IRV is a far better selection over Broda with a confidence interval of over 95\% and Contingent voting is again better than Broda with over 90\% confidence interval. There are cases where the difference does appear to be significant but the insights into these results are not merely restricted to significance in difference rather a guideline for better selection with all the possible options for voting than can be considered. And the research already establishes with previous results discussed in Figure. 3 and 4 how the conventional voting method or simply hard voting is not always the best option. GardensPoint dataset the results suggest that even though the difference is apparently substantial the identification of the better options such as IRV over Contingent and Plurality is a useful conclusion to be drawn. CrossSeasons dataset is another example of why random selection of a voting method is not the best idea given Contingent voting which is the ideal scenrio for a case like ESSEX3IN1 consistently under performs in comparison to other available methods such as Condorcet, IRV, Broda with a confidence interval ranging from even 90\% to 95\%.  Furthermore, in comparison to IRV and Broda, Plurality voting significantly performs better for this case with a z score of over 1.96. Similarly, Corridor dataset again does not stick to the pattern observed for ESSEX3IN1 and Contingent voting is not the better option to be selected over any of the other schemes. In fact Condorcet, IRV and Broda all are the better choice while IRV is better than plurality but Plurality is better than Broda. Lastly, 17Places the results vary alot as observed than Condorcet is better than Contingent, and so is IRV. Some cases the difference is not significant to consider while Plurality  is better to consider than some of the other choices. Overall, the heat-map is a useful reference point in navigating through the selection of a voting scheme in different ensemble cases to improve VPR performance. 

\section{Conclusion}
In this work we explore and implement the different universally used voting schemes that can be extended to an ensemble VPR set up to discover whether they provide any performance improvement over the conventional practises of voting. Furthermore we have also evaluated the performance difference via a McNemar's test like approach using a pairwise analysis to determine whether the difference in performance is statistically significant. With a selection of five employed and diverse voting methods which very commonly used among other fields of research we have tested these schemes on multiple datasets to provide the results presented. The selected methods tested include Plurality voting, Condorcet voting, Contingent Voting, Broda count and IRV voting. With several insights collected by the end, the first and foremost being that the employment of different voting schemes much like other fields does in fact produce very varied results when nothing just the voting method is different. With this we can confidently say that the selection of a voting method of a VPR ensemble set up is not a trivial task rather a process of careful selection. With different voting scheme standing out in performance for different types of datasets or variations in surroundings. Furthermore, among the tested voting methods the closest to the conventionally employed method is the Plurality method and interestingly the results showcase how the assumption that the common practise method might possibly be a good choice for all cases is not true at all. With the results we have collected it different conclusions for the selection process can be drawn given the requirement such as that a very basic ranking without any prerequisites of dataset type, variation or size would be that Condorcet Voting is overall consistent for most cases and produces substantial results followed by IRV and Contingent and then Broda. To further confirm our findings we have supported our results with a statistical analysis confirming the statistical significance of the results. These can be seen in the Z-score table in Table. 5 presenting z-score values that correspond to the confidence intervals to showcase how of our results presented in Figure. 3 are significant in difference. 

In the end, with a collection of insights into the voting system world it is safe to say that further research on the topic to improve and expand the utility of voting methods in the field of VPR is a relevant subject that holds alot more potential for more useful information.


\begin{thebibliography}{99}


\bibitem{c1} S. Lowry, N. Sunderhauf, P. Newman, J. Leonard, D. Cox, P. Corke and M. Milford. ”Visual Place Recognition: A Survey” IEEE Trans. Robot., 2016. 
\bibitem{c2} S. Garg, N. Suenderhauf and M. Milford, “Lost? appearance-invariant place recognition for opposite viewpoints using visual semantics,” in Proc. Robotics: Science and Systems, Jun. 2018
\bibitem{c3} F. Maffra, Z. Chen, and M. Chli, “Viewpoint-tolerant place recognition combining 2D and 3D information for UAV navigation,” in Proc. IEEE Int. Conf. Robot. Autom., 2018, pp. 2542–2549. 
\bibitem{c4} S. Hausler, A. Jacobson, and M. Milford, “Multi-process fusion: Visual place recognition using multiple image processing methods.” \textit{IEEE Robot.Autom. Lett.,} vol. 4, no. 2, 2019, pp. 1924–1931. 
\bibitem{c5} S. Hausler and M. Milford, "Hierarchical Multi-Process Fusion for Visual Place Recognition," \textit{In Proc. ICRA, IEEE} 2020.
\bibitem{c6} M. Waheed, M. J. Milford, K. Mcdonald-Maier and S. Ehsan, "Improving Visual Place Recognition Performance by Maximising Complementarity," in IEEE Robotics and Automation Letters, doi: 10.1109/LRA.2021.3088779.
\bibitem{c7} M. Waheed, M. Milford, K. McDonald-Maier and S. Ehsan. “SwitchHit: A Probabilistic, Complementarity-Based Switching System for Improved Visual Place Recognition in Changing Environments.” In Proc. IROS, IEEE, 2022
\bibitem{c8} B. Arcanjo, B. Ferrarini, M. Milford, K. D. McDonald-Maier, and S. Ehsan, “An efficient and scalable collection of fly-inspired voting units for visual place recognition in changing environments,” IEEE Robotics and Automation Letters, vol. 7, no. 2, pp. 2527–2534, 2022.
\bibitem{c9} M. Gehrig, E. Stumm, T. Hinzmann and R. Siegwart, "Visual Place Recognition with Probabilistic Voting." IEEE International Conference on Robotics and Automation (ICRA), 2017
\bibitem{c16----10}  A. C. Murillo and J. Kosecka, “Experiments in place recognition using gist panoramas,” in Proc. IEEE 12th Int. Conf. Comput. Vis. Workshops, 2009, pp. 2196–2203. 
\bibitem{c17----11} Y. Liu and H. Zhang, "Performance evaluation of whole-image descriptors in visual loop closure detection," in 2013 IEEE International Conference on Information and Automation (ICIA), 2013: IEEE, pp. 716-722.
\bibitem{c18-----12} N. Sünderhauf and P. Protzel, "Brief-gist-closing the loop by simple means," in 2011 IEEE/RSJ International Conference on Intelligent Robots and Systems, 2011: IEEE, pp. 1234-1241. 
\bibitem{c19-------13} Arandjelovic R, Gronat P, Torii A, Pajdla T, Sivic J (2016) Netvlad: Cnn architecture for weakly supervised place recognition. In: CVPR, pp 5297–5307
\bibitem{c20-------14} A. Khaliq, S. Ehsan, M. Milford and K. McDonald-Maier. “A holistic visual place recognition approach using lightweight cnns for significant viewpoint and appearance changes”, IEEE Trans. Robot., 2020

\bibitem{c21----15} J. Mount J, M. Milford, "2d visual place recognition for domestic service robots at night." IEEE International Conference on Robotics and Automation (ICRA), IEEE, pp 4822–4829, 2016
\bibitem{c10-----16} M. Milford, "Vision-based place recognition: how low can you go?," International Journal of Robotics Research, vol. Volume 32,Issue 7, pp 766–789, 2013.
\bibitem{c22------17} M. Zaffar, S. Ehsan, M. Milford, and K. D. McDonald-Maier, "Memorable maps: A framework for re-defining places in visual place recognition," IEEE Transactions on Intelligent Transportation Systems, 2018.
\bibitem{c23-------18} A. Glover, "A Day and night, left and right." Zenodo DOI 10.5281/zenodo.4590133, URL https://doi.org/10. 5281/zenodo.4590133 
\bibitem{c24------19} M. Larsson, E. Stenborg, L. Hammarstrand, M. Pollefeys, T. Sattler and F. Kahl. “A cross-season correspondence dataset for robust semantic segmentation”,In Proc. CVPR, pp 9532–9542, 2019.
\bibitem{c25------20} R. Sahdev and J. K. Tsotsos, "Indoor Place Recognition System for Localization of Mobile Robots," 2016 13th Conference on Computer and Robot Vision (CRV), Victoria, BC, Canada, 2016, pp. 53-60, doi: 10.1109/CRV.2016.38.
\bibitem{c26-------21} Z. Chen et al., "Deep learning features at scale for visual place recognition," in 2017 IEEE International Conference on Robotics and Automation (ICRA), 2017: IEEE, pp. 3223-3230, 2017.
\bibitem{c27------22} W. T. Freeman and M. Roth, “Orientation histograms for hand gesture
recognition,” in Proc. Int. Workshop Autom. Face Gesture Recognit., 1995, vol. 12, pp. 296–301.
\bibitem{c28-----23} N. Dalal and B. Triggs, “Histograms of oriented gradients for human detection,” in Proc. IEEE Comput. Soc. Conf. Comput. Vis. Pattern Recognit., 2005, vol. 1, pp. 886–893.
\bibitem{c29-----24} X. Han, Y. Zhong, L. Cao, and L. Zhang, “Pre-trained AlexNet architecture with pyramid pooling and supervision for high spatial resolution remote sensing image scene classification,” Remote Sens., vol. 9, no. 8, p. 848, 2017. [Online]. Available: https://www.mdpi.com/2072-4292/9/8/848
\bibitem{c30------25} N. Sunderhauf, S. Shirazi, F. Dayoub, Upcroft B, M. Milford. “On
the performance of convnet features for place recognition.”, In Proc.
IROS, IEEE, 2015, pp 4297–4304
\bibitem{c31----------26} Z. Chen et al., "Deep learning features at scale for visual place recognition," in 2017 IEEE International Conference on Robotics and Automation (ICRA), 2017: IEEE, pp. 3223-3230. 
\bibitem{c34------------------27} M. Zaffar, S. Ehsan, M. Milford, and K. McDonald-Maier, "CoHOG: A Light-Weight, Compute-Efficient, and Training-Free Visual Place Recognition Technique for Changing Environments," IEEE Robotics and Automation Letters, p. 99, 2020.
\bibitem{c35-----------------28} N. Merrill and G. Huang. “Lightweight Unsupervised Deep Loop Closure.”, In Proc. Robotics: science and systems (RSS), 2018.
for Changing Environments," IEEE Robotics and Automation Letters, p. 99, 2020.
\bibitem{29}S. Ehsan, A. Clark, A. Leonardis, A. Khaliq, M. Fasli, K. McDonald-Maier et al., “A
 generic framework for assessing the performance bounds of image feature detect
ors,” Remote Sensing, vol. 8, no. 11, p. 928, 2016
\bibitem{30}Q. McNemar, “Note on the sampling error of the difference between correlated
 proportions or percentages,” Psychometrika, vol. 12, no. 2, pp. 153–157, 1947.
\bibitem{31}J. L. Fleiss, B. Levin, and M. C. Paik, Statistical methods for rates and proportions. John Wiley \& Sons, 2013.




 







\end{thebibliography}
\end{document}